\documentclass[runningheads]{llncs}
\usepackage{graphicx}
\usepackage{amsmath,amssymb} 
\usepackage{color}
\usepackage[width=122mm,left=12mm,paperwidth=146mm,height=193mm,top=12mm,paperheight=217mm]{geometry}
\begin{document}
\pagestyle{headings}
\mainmatter

\title{Pose-Based Two-Stream Relational Networks for Action Recognition in Videos} 

\titlerunning{Pose-Based Two-Stream Relational Networks for Action Recognition in Videos}

\authorrunning{Wei Wang, Jinjin Zhang, Chenyang Si, Liang Wang}

\author{Wei Wang$^{1,3}$ \and Jinjin Zhang$^{1}$ \and Chenyang Si$^{1,3}$ \and Liang Wang$^{1,2,3}$   \\
    $^1$Center for Research on Intelligent Perception and Computing (CRIPAC), \\
    National Laboratory of Pattern Recognition (NLPR)\\
    $^2$Center for Excellence in Brain Science and Intelligence Technology (CEBSIT),\\
    Institute of Automation, Chinese Academy of Sciences (CASIA)\\
    $^3$University of Chinese Academy of Sciences (UCAS)\\
    {\tt\small \{wangwei, wangliang\}@nlpr.ia.ac.cn, \{jinjin.zhang, chenyang.si\}@cripac.ia.ac.cn }
}
\institute{}

\maketitle
\begin{abstract}
Recently, pose-based action recognition has gained more and more attention due to the better performance compared with traditional appearance-based methods. However, there still exist two problems to be further solved. First, existing pose-based methods generally recognize human actions with captured 3D human poses which are very difficult to obtain in real scenarios. Second, few pose-based methods model the action-related objects in recognizing human-object interaction actions in which objects play an important role. To solve the problems above, we propose a pose-based two-stream relational network (PSRN) for action recognition. In PSRN, one stream models the temporal dynamics of the targeted 2D human pose sequences which are directly extracted from raw videos, and the other stream models the action-related objects from a randomly sampled video frame. Most importantly, instead of fusing two-streams in the class score layer as before, we propose a pose-object relational network to model the relationship between human poses and action-related objects. We evaluate the proposed PSRN on two challenging benchmarks, i.e., Sub-JHMDB and PennAction. Experimental results show that our PSRN obtains the state-the-of-art performance on Sub-JHMDB (80.2\%) and PennAction (98.1\%). Our work opens a new door to action recognition by combining 2D human pose extracted from raw video and image appearance.

\end{abstract}

\section{Introduction}

Recognition of human actions is a very important task in computer vision, which has a wide range of applications, e.g., intelligent video surveillance, robot vision, human-computer interaction, and so on. Action recognition is also a very challenging task, which has to analyze human motion information and understand its temporal characteristics. Moreover in most cases, the objects interacted with humans provide an important clue for action recognition.

For a long time, learning and extracting effective spatio-temporal features for videos is the mainstream direction of action recognition, i.e., the well-known spatio-temporal interest points (STIPs) \cite{STIP} and improved dense trajectories (iDT) \cite{IDT}. Recently with the advent of deep learning, deep convolutional networks for action recognition have received significant attention due to the large improvement on recognition accuracy. Tran et al. \cite{3DCNN} extend traditional 2D convolutional network to its 3D version to learn spatiotemporal features for video action recognition. Simonyan et al. \cite{two_stream} propose the first two-stream ConvNet architecture for both temporal optical flow and spatial appearance, which is followed by several improved architectures \cite{two_stream_feich17}\cite{TSN}. Although these two-stream approaches achieve better performance on the benchmarks, i.e., UCF101 \cite{UCF101}, the optical flow in temporal stream generally demands a high computation cost, and the two streams lack of a principled fusion strategy.

Actually, 2D/3D human poses as the trajectories of skeleton joints are more effective representations for characterizing the dynamics of human actions if comparing with optical flow. Analyzing pose/skeleton sequence is another approach for action recognition. Du et al. \cite{HRNN} propose a hierarchical recurrent neural network for skeleton based action recognition by dividing human skeleton into five parts according to human physical structure. To extract discriminative skeleton features, Song et al. \cite{spatialtemporalattention} propose a spatio-temporal attention model. Pose-based methods have achieved great success on the datasets, i.e., NTU RGB+D \cite{ntu}. However, there exist two urgent problems to be solved. First, instead of starting from raw video, most of the pose-based methods recognize actions directly from captured 3D human poses which are very difficult to obtain in real scenarios. Second, existing pose-based methods generally ignore to exploit spatial appearance which contains the objects associated with particular actions. Although Du et al. \cite{RPAN} combine pose and appearance for action recognition, they just introduce pose-based attention to learn human-part appearance features at each time, and ignore to model the action-related objects as well.

From the analysis above, fusing both human poses and action-related objects becomes a natural and reasonable choice to recognize human actions. Recent progress in realtime 2D pose estimation from image and video, i.e., openpose \cite{cao2017realtime}, has made action recognition from estimated poses plausible. The work on relational reasoning \cite{relational_network} inspires us to provide a principled way to model the relationship between human poses and action-related objectss.

\begin{figure}[t]
\centering \includegraphics[width=1.0\linewidth]{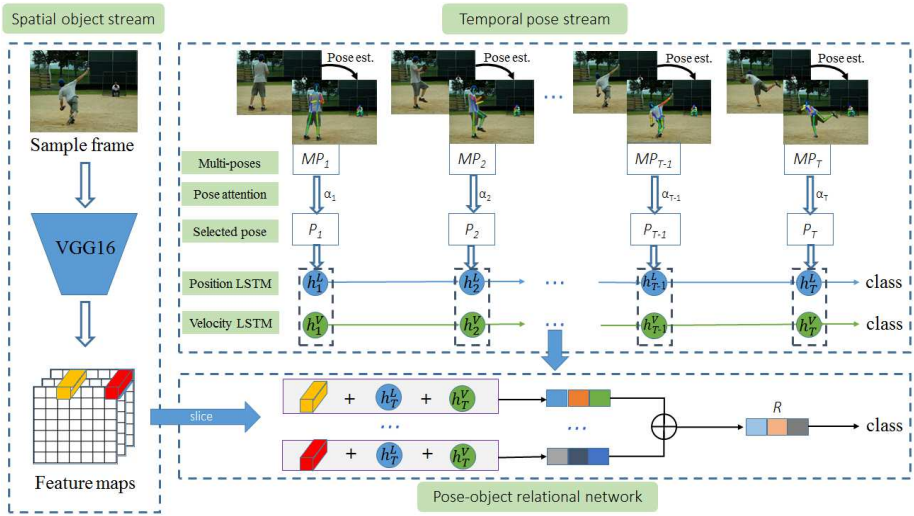}
   	\caption{The architecture of the proposed pose-based two-stream relational network (PSRN). It contains a temporal pose stream, a spatial object stream and a pose-object relational network. The pose stream models the temporal dynamics of the targeted 2D human pose sequences, and the object stream extracts visual feature maps from a randomly sampled video frame to represent the action-related objects. Pose-object relational network provides a principled two-stream fusion strategy by modelling the relationship between human poses and action-related objects.}
   	\label{fig.psrn}
\end{figure}

In this paper, we propose a pose-based two-stream relational network (PSRN) for action recognition. Fig. \ref{fig.psrn} shows the architecture of PSRN which contains a temporal pose stream, a spatial object stream and a pose-object relational network. In the pose stream, we first extract the multi-person 2D poses of video frames with an improved openpose. Due to the varying number of persons in videos, we assume there is only one person performing the actions, and further select the targeted pose with an attention mechanism from the multi-person poses. To better represent human poses, we not only utilize the skeleton position but also the skeleton velocity, both of which are modelled with LSTM-RNNs to characterize the pose dynamics. The last hidden representations in these two LSTM-RNNs, i.e., $h^L_T$ and $h^V_T$, will be used for later pose-object relational modelling.
The object stream extracts the features maps of a randomly sampled video frame by a VGG16 network to represent the action-related objects.
In the pose-object relational network, instead of fusing two-streams in the class score layer as before, we take each column of the object feature maps as an object, and combine all the columns with pose hidden representations $h^L_T$ and $h^V_T$ to obtain a relation feature $R$. Finally, we supervise PSRN with three classification losses on the pose hidden representations $h^L_T$, $h^V_T$ and relation feature $R$, respectively. This kind of multi-loss objective makes our model learn better.

We perform experiments on two challenging action recognition datasets,
namely PennAction \cite{pennaction} and Sub-JHMDB \cite{sub-jhmdb}, to verify the effectiveness of our model.
Experimental results show that the proposed PSRN outperforms the state-of-the-art on these datasets.

The main contributions of this paper are summarized as follows:
\begin{enumerate}
  \item We propose a new two-stream architecture for both 2D pose sequence and image appearance, which provides effective representations for the temporal dynamics of human actions and action-related objects.
  \item We propose a principled strategy to fusion the two-stream networks by modelling the relationship between human pose and action-related objects with a relational network.
  \item The proposed pose-based two-stream relational network (PSRN) achieves the best results on two challenging benchmarks, which verifies its effectiveness.
  \item Our work opens a new door to action recognition by utilizing 2D human poses directly extracted from raw videos, which no longer needs to input captured 3D human poses or expensive optical flows.
\end{enumerate}

The remainder of this paper is organized as follows. In Section \ref{sect:relatedwork}, we introduce the related work on action recognition and relational modelling. In Section \ref{sect:PSRN}, we illustrate the details of the proposed PSRN. Experimental results are presented in Section \ref{sect:experiments}. Finally, we conclude and discuss the paper in Section \ref{sect:conclusion}.

\section{Related Works}\label{sect:relatedwork}

In this section, we briefly review the existing literature that closely relates to the proposed pose-based two-stream relational network, including 2D human pose estimation, pose-based action recognition, two-stream action recognition and relational modelling.

\emph{2D human pose estimation} \hspace{3mm} There exist two categories of 2D human pose estimation methods: single-person and multi-person. For single-person pose estimation, Toshev et al. \cite{DeepPose} propose an architecture called DeepPose which is the first pose estimation method using deep networks. Beyond using ConvNet for joint estimation, some work builds grapical model to learn spatial relationships between joints \cite{graphical_pose}. Recently, several well-designed deep networks have achieved impressive results \cite{hourglass}. For multi-person pose estimation, Cao et al. \cite{cao2017realtime} exploit part affinity fields (PAF) to associate body parts with individuals in a bottom-up way. Some top-down approaches consist of two stages of person detector and single-person pose estimation for each detection \cite{multiposewild}. He et al. propose an end-to-end solution called Mask R-CNN by extending Faster R-CNN \cite{mask}. In this paper, we finetune the PAF proposed in \cite{cao2017realtime} on a larger skeleton detection dataset and obtain much more accurate poses on the experimental action datasets.

\emph{Pose-based action recognition} \hspace{3mm} Recognition of human actions based on pose sequence has received much more attention, due to its effective representation on action dynamics. Previously, various handcrafted features, i.e., the relative positions of joints and the covariance matrix of joint locations over time \cite{hod}, are proposed as discriminative descriptors for pose sequence. Recently, large amounts of approaches using recurrent neural network (RNN) are proposed to model the temporal dynamics of pose sequence. Du et al. \cite{HRNN} propose an end-to-end hierarchical RNN for skeleton based action recognition by dividing the human skeleton into five parts according to human physical structure. Song et al. \cite{spatialtemporalattention} propose an end-to-end spatial and temporal attention model for action recognition. Zhang et al. \cite{viewadaptive} propose a view adaptive model which can adapt to the most suitable viewpoints for action recognition. Although pose-based methods have achieved great success on several benchmarks, most of them recognize actions directly from captured 3D human pose which is very difficult to obtain in real scenarios. In this paper, we model the action dynamics with 2D human poses which are extracted from raw videos.

\emph{Two-stream action recognition} \hspace{3mm} Inspired by the two-stream hypothesis for perception and action \cite{perception}, Simonyan et al. \cite{two_stream} propose a two-stream ConvNet architecture to process temporal motion information and spatial appearance in parallel, and then fuse the classification scores of these two streams. Several extensions to that work have been proposed to explore the two-stream architectures. Feichtenhofter et al. \cite{two_stream_feich17} investigate convolutional spatial fusion and temporal fusion, i.e., bilinear fusion and 3D pooling, especially the multiplicative interactions of spacetime features. For long-range temporal structure modeling, Wang et al. \cite{TSN} propose a temporal segment network for video-based action recognition, which combines a sparse temporal sampling strategy and video-level supervision. It is well known that the optical flow used in temporal stream generally demands a high computation cost to obtain reasonable accuracy. In this paper, we use 2D pose sequence to represent motion information, which is much more natural and reasonable than optical flow. In addition, we propose to fuse two streams with a novel pose-object relational network.

\emph{Relational modelling} \hspace{3mm} Graph-based methods inherently support relation-centric computation, i.e., graph neural networks \cite{GNN} and interaction networks \cite{interaction}. Santoro et al. \cite{relational_network} recently propose a simple plug-and-play general solution to relational reasoning. To handle complex tasks which requires an order of magnitude more steps of relational reasoning, i.e., Sudoku puzzle, Palm et al. \cite{recurrentrelation} propose a recurrent relational network. In this paper, we follow the idea of \cite{relational_network} and propose a pose-object relational network to model the relationship between human pose and action-related objects.

\section{Pose-Based Two-Stream Relational Network}\label{sect:PSRN}

In this section, we explain the proposed pose-based two-stream relational network in detail. First, we introduce the procedure of estimating 2D human poses from raw videos. Then, we describe the temporal pose stream and spatial object stream, respectively. Next, we illustrate the pose-object relational network. Finally, we give the details of learning PSRN.

\subsection{2D Human Pose Estimation}\label{subsect:poseEstimation}

\begin{figure}[t]
\begin{center}
\includegraphics[width=1 \linewidth]{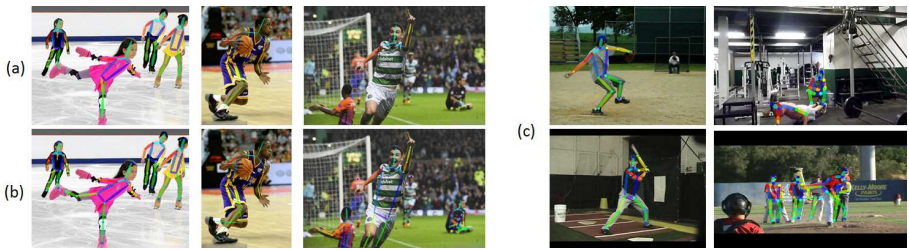}
\end{center}
\caption{Results of 2D human pose estimation. (a) and (b) The testing results from AI challenge dataset by \cite{cao2017realtime} and our implementation. Our retrained model performs much better than the implementation in \cite{cao2017realtime}. (c) Several kinds of failure cases of our retrained model on action videos, i.e., keypoint missing on body part or whole body, keypoints detected on action-related objects.} \label{fig:pose_est}
\end{figure}

This work does not aim to propose a new pose estimation method, so we choose the approach presented in \cite{cao2017realtime} to estimate the multi-person poses of action videos. This approach also called PAF proposes a part affinity field to associate body parts in a bottom-up way, which maintains high accuracy and achieves realtime performance. In our experiments, we retrain PAF on a larger AI challenge dataset\footnote{https://challenger.ai/competition/keypoint/subject}, which has 210,000 annotated images in the training set. Compared to the 18 keypoints in COCO 2016 keypoints challenge dataset used in \cite{cao2017realtime}, this dataset annotates only 14 keypoints for each person, which ignores 5 keypoints on two eyes, two ears and one nose, and adds a new keypoint on top of the head. To obtain more accurate detection results on action datasets, we replace original VGG16 with a powerful inception-resnet-V2 in the front-end of PAF. The retrained model performs much better than its original implementation with the mAP of 0.52 to 0.48 on the AI challenge dataset. Fig. \ref{fig:pose_est}(a) and (b) shows several testing results from AI challenge dataset by using \cite{cao2017realtime} and our implementation, respectively. Although our retrained model improves the accuracy of keypoint detection, there still exist several kinds of failures on the action video dataset as shown in Fig. \ref{fig:pose_est}(c), i.e., keypoint missing on body part or whole body, keypoints detected on action-related objects. These failures pose great challenges for further action recognition. However, from the experimental results in Section \ref{sect:experiments}, we can see that the proposed PSRN still achieves better results.
\subsection{Temporal Pose Stream}\label{subsect:posestream}
\begin{figure}[t]
\begin{center}
\includegraphics[width=0.9\linewidth]{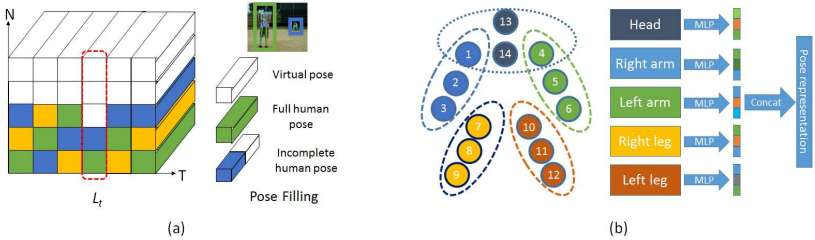}
\end{center}
\caption{(a) Pose filling to obtain consistent and complete pose data. We add virtual human poses and  complement incomplete poses with $(0,0)$ keypoints. (b) Similar to \cite{HRNN}, the 14 keypoints of a human pose are divided into five parts annotated by ellipse boxes according to human physical structure. The representation of a human pose is the concatenation of the five part representations.} \label{fig:pose_rep}
\end{figure}

Considering that there exist varying number of persons in the action videos, and some human poses extracted in Section \ref{subsect:poseEstimation} may be incomplete, we need to perform pose filling to obtain consistent and complete pose data. We count the number of persons in each frame and choose the max number $N$ as the common person number of all the videos. If the keypoints of some body part are missing or the number of detected persons is smaller than $N$, we need to fill keypoint $(0,0)$ into the missing parts/persons so that each frame has a complete representation for $N$ persons. Fig. \ref{fig:pose_rep}(a) illustrates the procedure of pose filling, which adds virtual human poses with all $(0,0)$ keypoints and complements part pose.

For each of the $N$ human poses, Fig. \ref{fig:pose_rep}(b) shows its 14 keypoints which are divided into five parts annotated by ellipse boxes. Each part is initially represented by the concatenation of keypoint positions and then transformed into a $K$-dim vector by a multilayer peceptron (MLP). The representation of a human pose is the concatenation of these five $K$-dim vectors.

Actually, most of the human actions occur on a specific person, so that it is very difficult to obtain a compact representation for action dynamics if directly concatenating all the human poses at each frame. We propose to select the targeted pose from all the detected poses by virtue of attention mechanism.

At each time-step $t$, we have obtained $N$ $5K$-dim  vectors for all the human poses, which are denoted as
\begin{eqnarray}
\label{formu:L}
 L_t= [L_{t,1}, ..., L_{t,N}], \hspace{9mm} L_{t,i}\in \mathbb{R}^{5K}
\end{eqnarray}

We adopt the soft attention mechanism introduced in Xu et al. \cite{attendcaption}, which is based on LSTM implementation:
\begin{eqnarray}
\label{eqn:lstm}
\left( \begin{array}{l}
{i_t}\\
{f_t}\\
{o_t}\\
{g_t}
\end{array} \right) &=& \left( \begin{array}{l}
\sigma \\
\sigma \\
\sigma \\
\tanh
\end{array} \right)M\left( \begin{array}{l}
{h_{t - 1}}\\
{l_t}
\end{array} \right)\\
{c_t} &=& {f_t} \odot {c_{t - 1}} + {i_t} \odot {g_{t - 1}}\\
{h_t} &=& {o_t} \odot \tanh ({c_t})
\end{eqnarray}
where $i_t$ is the input gate, $f_t$ is the forget gate, $o_t$ is the output gate, $g_t$ is the intermediate memory state, $c_t$ is the updated memory state and $h_t$ is the hidden state. $l_t$ represents the input to the LSTM at time-step $t$, which will be explained later. $M_{5K+d, 4d}$ is the trainable parameters of an affine transformation, here $d$ is the dimension of all the gates and states. $\sigma$ and $\odot$ are the sigmoid activation and element-wise multiplication.

The soft attention computes a positive weight to measure the importance of
each pose representation $L_{t,i}$ based on previous hidden state $h_{t-1}$. The weight $\alpha_{t,i}$ is generally computed by a multilayer perceptron $f_{att}$ which is followed by a softmax operation:
\begin{eqnarray}
e_{t,i} &=& f_{att}(L_{t,i}, h_{t-1})\\
\alpha_{t,i} &=& \frac{\exp(e_{t,i}))}{\sum_k{\exp(e_{t,k})}} \nonumber
\end{eqnarray}

\begin{figure}[t]
\begin{center}
\includegraphics[width=1.0\linewidth]{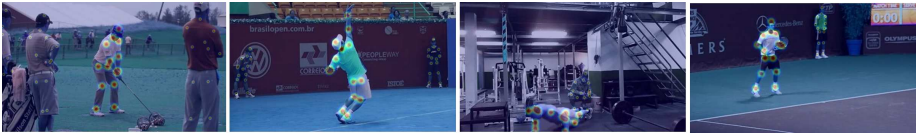}
\end{center}
\caption{The learned attention weights for multiple human poses. The size of the keypoint area denotes the weight value. We can see that the targeted human pose has larger weights than the non-targeted human poses. } \label{fig:saliency}
\end{figure}

Fig. \ref{fig:saliency} shows the learned attention weights for several action videos. The size of keypoint area denotes the weight value. We can see that the targeted human pose has larger weights than the non-targeted poses.

Once the attention weights have been computed, the selected pose representation is a summation of the weighted pose representations
\begin{eqnarray}
l_t = \sum_i{\alpha_{t,i}L_{t,i}}
\end{eqnarray}

As shown in Eqn. \ref{eqn:lstm}, $l_t$ is the selected compact pose representation which is input to the LSTM.

Up to the last time-step $T$, the hidden representation $h_T$ of the LSTM can be used to classify actions and build the pose-object relational network later. For clarity, we replace $h_T$ with $h^L_T$ to denote the representation of pose position.

To further explore the temporal dynamics of pose sequence, we compute the pose velocity $V_t$ by differencing the selected pose representations $l_t$ and $l_{t+1}$
\begin{eqnarray}
V_t = l_{t+1} - l_{t}
\end{eqnarray}

The pose velocity sequence $\{V_t\}_{t=1}^{T}$ can be modelled by another LSTM. The output $h_{T}^V$ of this LSTM can also be used to classify actions and build the pose-object relational network later.

\subsection{Spatial Object Stream}\label{subsect:objectstream}

Spatial appearance provides a useful clue for action recognition since some actions are tightly associated with particular objects. In the literature, there are large amounts of work on action classification from still images \cite{stillaction}.

In this work, we randomly sample a video frame and  extract its feature maps $X$ ($7\times7\times512$) with a VGG16 network. Each 512-dim column vector $x_i$ is considered as an object representation. We send the 49 object representations into the next module for pose-object relational modelling.

\subsection{Pose-Object Relational Network}\label{subsect:relation}

After obtaining the pose representations $h_T^L$ and $h_T^V$ from temporal pose stream, and the object representations $X\in\{x_i\}_{i=1}^{49}$ from spatial object stream, we model the relationship between human poses and action-related objects, which provides a principled two-stream fusion strategy. Inspired by \cite{relational_network} performing relational reasoning on question answering tasks, we define the pose-object relation as a composite function below:
\begin{eqnarray}
R = RN(P,O) = {f_\phi }(\sum\limits_i {{g_\theta }(h_T^L,h_T^V,{x_i})} )
\end{eqnarray}
where $R$ is the relation representation, $f_\phi$ and $g_\theta$ are simple multilayer perceptrons (MLP) with parameters $\phi$ and $\theta$, respectively.

\subsection{Learning PSRN}\label{subsect:implementdetail}

The proposed PSRN is a principled framework for video-based action recognition. If aiming to achieve better performance, the design of network architecture and the procedure of network training should be taken more care of. In this section, we introduce the details of the network architecture and network training.

{\bf Network Architecture} \hspace{2mm} To further improve the performance, we enhance the temporal pose stream by replacing original two unidirectional LSTMs (uni-LSTMs) with two bidirectional LSTMs (bi-LSTMs) on both pose position and pose velocity. For modelling long-term dependency on pose sequences with LSTMs, we adopt the lookback structure which looks at the outputs from the last $n$ steps when generating the output for the current step. Compared to other two-stream architectures, i.e., \cite{TSN} and \cite{RPAN} which adopt very deep convolutional networks for optical flow and image appearance, our two-stream networks which only contain a VGG16 network on images and two bidirectional LSTMs on pose sequences are not very complicated.

{\bf Network Training} \hspace{2mm} Considering that both of the two-stream networks can be used to perform recognition task, we supervise our model with the following total loss:
\begin{eqnarray}
\mathcal{L}_{total} = \mathcal{L}_{pos} + \mathcal{L}_{vel} + \mathcal{L}_{rel} + \lambda_{\theta}||\theta||_2
\end{eqnarray}
where $\mathcal{L}_{pos}$, $\mathcal{L}_{vel}$ and $\mathcal{L}_{rel}$ are three cross-entropy losses corresponding to pose position representation $h_T^L$, pose velocity representation $h_T^L$ and relation representation $R$, respectively. $||\theta||_2$ is the weight decay regularization for all the model parameters, and $\lambda_{\theta}$ is the weight decaying coefficient which is set to $0.00004$ in the experiments. It should be noted that the same coefficients are set to the three losses, because there is no obvious improvement using different coefficients in the experiments.

Due to the similar form of these three cross-entropy losses, we take $\mathcal{L}_{pos}$ as an example
\begin{eqnarray}
\mathcal{L}_{pos} = -\sum_{m=1}^M\sum_{c=1}^Cy_{m,c}\log\hat{y}_{m,c}^{pos}
\end{eqnarray}
there are $M$ action videos and $C$ classes of human actions. $y_{m,c}$ is the groundtruth label while $\hat{y}_{m,c}^{pos}$ is its prediction when using pose positions.

During training, we use the Adam optimizer \cite{adam} with $\beta_1$ = 0.9, $\beta_2$ = 0.999. Practically, it is difficult for the proposed PSRN to converge to a good solution when training the whole network from the scratch. We adopt a three-stage training strategy to provide better initializations for the final end-to-end training. In the first stage, we train the temporal pose stream with the losses $\mathcal{L}_{pos}$ and $\mathcal{L}_{vel}$. The initial learning rate is set to 0.0001 and reduced to its half after 78,000 iterations. In the second stage, we fix the learned parameters of the temporal pose stream, and only train the spatial object stream and pose-object relational network with the loss $\mathcal{L}_{rel}$. The VGG16 network in spatial object stream is initialized with a widely-used model pretrained on ImageNet dataset. For a better convergence, we vary the learning rate in a similar way to the warmup strategy introduced in \cite{attentionallyouneed}, which increases the learning rate exponentially for the first warmpup steps (2000) from 1e-6 to 1e-4, and then decreases it to its half after 28,000 iterations. In the third stage, we start to train the whole network which is initialized with previous learned parameters with the total loss $\mathcal{L}_{total}$.

\section{Experiments}\label{sect:experiments}

In this section, we first introduce the experimental datasets and implementation details. Then, we compare the proposed PSRN with several state-of-the-art methods. Finally, we perform model analysis by evaluating the key model components and analyzing the confusion matrix of the classification results on PennAction.

\subsection{Datasets}

We evaluate our pose-based two-stream relational network (PSRN) on two benchmarks in pose-related action recognition, i.e., Sub-JHMDB \cite{sub-jhmdb} and PennAction \cite{pennaction}. These two datasets are very challenging due to the richer variation in terms of appearance and dynamics. It should be noted that although the full body human joints are annotated for the videos in these datasets, we do not use them during model training and testing. In addition, we use the evaluation protocol introduced in \cite{RPAN} to report classification accuracy for both datasets.

{\bf Sub-JHMDB:}\hspace{1mm} It is a subset of JHMDB, which contains 316 clips distributed over 12 action categories, i.e., swing\_baseball, climp\_stairs, kick\_ball, walk, jump, pullup, push, pick, catch, run, shoot\_ball and golf. Each clip contains between 15 and
40 frames of size 320 $\times$ 240. There are 3 train/test splits. Similar to \cite{RPAN}, we compare the average accuracy on these three splits to the state-of-the-art.

{\bf PennAction:}\hspace{1mm} It consists of 2326 challenging
consumer videos distributed over 15 action categories, i.e., baseball\_pitch, baseball\_ swing, bench\_press, bowling, clean\_and\_jerk, golf\_swing, jump\_rope, jumping\_jacks, pullup, pushup, situp, squat, strum\_guitar, tennis\_forehand and tennis\_serve. This dataset has rich annotations which consist of action class labels, 2D keypoint positions and their corresponding visibilities, and camera viewpoints.

\subsection{Implementation Details}

If no otherwise specified, we perform our PSRN on the two datasets with the same implementation details. The sampled video frame is randomly cropped and resized to 224 $\times$ 224 for the VGG16 network to extract spatial feature maps (7 $\times$ 7 $\times$ 512), and the videos are transformed to multiple scales for the retrained PAF to estimate human poses (14 keypoints). For convenience, we rescale the pose positions between 0 and 1. As introduced in Section \ref{subsect:posestream}, human poses are divided into five parts which have the following position dimensions: head (8-dim), left/right arm (6-dim) and left/right leg (6-dim). The body part positions are initially transformed into 100-dim representations with MLPs. Then the representation of human pose is the concatenated 500-dim vector which is input to the LSTM-based pose stream. Both of the position and velocity LSTMs have 512-dim hidden representation. The four-layer MLP for $g_{\theta}$ and the two-layer MLP for $f_{\theta}$ in pose-object relational network all consist of 512 units per layer with ReLU non-linearities. During both training and testing stages, we randomly sample 10 frames from each video as the input to the temporal pose stream, and resample one frame from these 10 frames as the input to the spatial object stream. The lookback steps in our LSTM implementations are setting to 5. We implement our PSRN by Tensorflow\footnote{https://www.tensorflow.org}.

\subsection{Experimental Results}

To evaluate the performance of the proposed PSRN, we compare it with the recent state-of-the-art approaches in pose-based action recognition on Sub-JHMDB and PennAction. Some of them belongs to the hand-crafted approaches (Action Bank \cite{detailedaction}, MST \cite{zhu2014}, AOG \cite{zhu2015} and Hierarchical \cite{hierarchical}), while the others are exploiting deep networks (P-CNN \cite{p-cnn}, JDD \cite{jdd}, Pose+idt-fv \cite{dt-fv}, RPAN \cite{RPAN}). The experimental results are shown in Table \ref{table:stat}. We can see that our PSRN with only pose stream achieves the comparable results with some approaches. This is mainly credited to the attention mechanism and bidirectional LSTMs on both pose positions and pose velocities. Our PSRN with two-stream relational network achieves the best results and outperforms all the other methods on both datasets. This further verifies the role of action-related objects in action recognition and the effectiveness of our pose-object relational modelling.

\setlength{\tabcolsep}{4pt}
\begin{table}
\begin{center}
\caption{Comparison with the state-of-the-art approaches on Sub-JHMDB (average
over three splits) and PennAction. Our PSRN with two-stream relational network achieves the best performance.}
\label{table:stat}
\begin{tabular}{llll}
\hline\noalign{\smallskip}
State-of-the-art & Year & Sub-JHMDB & PennAction\\
\noalign{\smallskip}
\hline
\noalign{\smallskip}
Action Bank \cite{detailedaction} & 2013 & - & 83.9\\
MST \cite{zhu2014}& 2014 & 45.3 & 74.0\\
AOG \cite{zhu2015}& 2015 & 61.2  & 85.5\\
P-CNN \cite{p-cnn}& 2015 & 66.8 & -\\
Hierarchical \cite{hierarchical} & 2016 & 77.5 & -\\
JDD \cite{jdd}& 2016 & 77.7 & 87.4\\
Pose+ idt-fv \cite{dt-fv}& 2017 & 74.6 & 92.9\\
RPAN \cite{RPAN}& 2017 & 78.6 & 97.4\\
\hline
Our PSRN (only pose stream) &   & {\bf 71.7} & {\bf 95.9}\\
Our PSRN (two-stream) &   & {\bf 80.2} & {\bf 98.1}\\
\hline
\end{tabular}
\end{center}
\end{table}
\setlength{\tabcolsep}{1.4pt}

\subsection{Model Analysis}

To understand the properties of the proposed PSRN, we first evaluate the effectiveness of several key model components on Sub-JHMDB (the second split) and PennAction, i.e., attention mechanism on pose selection, bi-LSTM on pose sequence modelling and two-stream relational network. Then we analyze the confusion matrix of the classification results on PennAction.

{\bf Architecture Analysis:}\hspace{1mm} Table \ref{table:sub_model_analysis} and \ref{table:penn_model_analysis} show the classification accuracies of several PSRN variants, which either replace bidirectional LSTM with unidirectional LSTM (uni-LSTM+Attention) or ignore the attention mechanism to select targeted poses (bi-LSTM). Moreover, we give the accuracies from pose position, pose velocity, pose stream fusion (position and velocity), and the proposed two-stream relation fusion (relational network). We can see that under the same stream settings, our PSRN with bi-LSTMs and attention achieves better performance than the other two variants, which verifies the effectiveness of bi-LSTMs and attention mechanism. With the same network architecture, our two-stream relation fusion outperforms other streams, which verifies the effectiveness of the proposed pose-object relational network.

\setlength{\tabcolsep}{4pt}
\begin{table}
\begin{center}
\caption{Evaluating the effectiveness of bi-LSTMs, attention mechanism and two-stream relational network on the second split of Sub-JHMDB.}
\label{table:sub_model_analysis}
\begin{tabular}{llll}
\hline\noalign{\smallskip}
Stream & uni-LSTM+Attention & bi-LSTM & bi-LSTM+Attention\\
\noalign{\smallskip}
\hline
\noalign{\smallskip}
Pose Position  & 65.3 & 53.8 & 70.0\\
Pose Velocity & 66.3 & 57.5 & 71.2 \\
Pose Stream Fusion & 68.8 & 60.0 & 73.8 \\
Two-stream Fusion & 77.5 & 65.0 & 83.7\\
\hline
\end{tabular}
\end{center}
\end{table}
\setlength{\tabcolsep}{1.4pt}

\setlength{\tabcolsep}{4pt}
\begin{table}
\begin{center}
\caption{Evaluating the effectiveness of bi-LSTMs, attention mechanism and two-stream relational network on PennAction.}
\label{table:penn_model_analysis}
\begin{tabular}{llll}
\hline\noalign{\smallskip}
Stream & uni-LSTM+Attention & bi-LSTM & bi-LSTM+Attention\\
\noalign{\smallskip}
\hline
\noalign{\smallskip}
Pose Position  & 93.9 & 84.8 & 94.0\\
Pose Velocity & 94.7 & 90.9 & 95.4 \\
Pose Stream Fusion & 95.4 & 92.3 & 95.9 \\
Two-stream Fusion & 96.9 & 94.9 & 98.1 \\
\hline
\end{tabular}
\end{center}
\end{table}
\setlength{\tabcolsep}{1.4pt}

\begin{figure}[h]
\begin{center}
\includegraphics[width=1.0\linewidth]{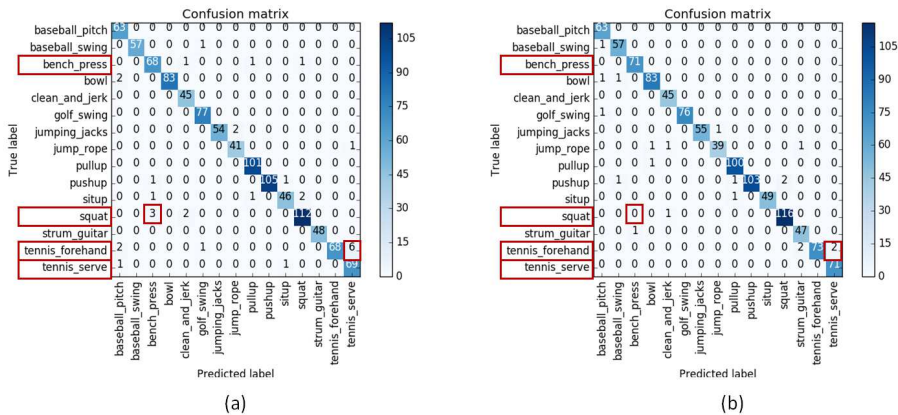}
\end{center}
\caption{Confusion matrix comparison on PennAction. (a) RPAN \cite{RPAN} and (b) Our PSRN. The values in the matrix are the number of test videos. Our PSRN misclassifies much less videos than RPAN with 20 vs. 31. The improvements are mainly from 4 human actions, i.e., bench\_press (3), squat (4), tennis\_forehand (5) and tennis\_serve (2). Moreover, RPAN often misclassifies tennis\_forehand as tennis\_serve and squat as bench\_press, while our PSRN reduce these confusions by 6-2=4 and 3, respectively. The comparison results show the advantages of our pose-based two-stream relational network.} \label{fig:confusion}
\end{figure}

{\bf Confusion Matrix:}\hspace{1mm} Furthermore, we analyze the classification results with confusion matrix and compare it with RPAN \cite{RPAN} on PennAction.  Fig. \ref{fig:confusion} (a) and (b) show the confusion matrix of RPAN and ours, respectively. Our PSRN with 20 misclassified videos improves RPAN with 31 misclassified videos. The improvements mainly come from 4 human actions, i.e., bench\_press (3), squat (4), tennis\_forehand (5) and tennis\_serve (2). We can see that all of these four classes have action-related objects, and our model can model these objects and their relations to the corresponding actions by virtue of the proposed pose-object relational network. Moreover, RPAN often misclassifies tennis\_forehand as tennis\_serve and squat as bench\_press, while our PSRN reduces these confusions although these action pairs have very similar spatial appearance. This verifies the effectiveness of our pose-based temporal streams which can model and distinguish action dynamics better.

\section{Conclusion and Future Work}\label{sect:conclusion}

In this paper, we propose a pose-based two-stream relational
network (PSRN) for action recognition. One stream models the temporal dynamics of the targeted 2D human pose sequences, while the other stream represents the spatial objects. Moreover, we propose a principled way to fuse these two streams. The proposed PSRN achieves the best performance on two challenging benchmarks.

More efforts should be made to further improve our work. Firstly, our current relational model can not explicitly tell us which objects are related to human poses/actions. We can propose a selective relational network which assigns weights for all the objects and further selects the most relevant object for pose-object relational modelling. Secondly, this work focuses on single-person action recognition based on the hypothesis that most of the human actions occur on a specific person, which should be extended to handle multi-person action recognition. Thirdly, 2D pose estimation is independent from the other parts of PSRN in this version. To perform an end-to-end learning and achieve optimal performance, we need to seamlessly integrate the pose estimation into our model.



\bibliographystyle{splncs}
\bibliography{egbib}
\end{document}